\documentclass[letterpaper, 10 pt, conference]{ieeeconf}  %
\usepackage{cite}  %
\usepackage{pifont}
\usepackage{cite}
\usepackage{amsmath} %
\usepackage{amsmath,amssymb,amsfonts}
\usepackage{algorithmic}
\usepackage{graphicx}
\usepackage{textcomp}
\usepackage{xcolor}
\usepackage{url}
\usepackage{gensymb}
\usepackage{siunitx}
\usepackage{censor}
\usepackage{comment}
\usepackage{balance}
\usepackage{float}
\usepackage[ruled,vlined]{algorithm2e} %

\def \standardreducedwidth{0.89}

\IEEEoverridecommandlockouts                              %

\title{\LARGE \bf
DeRP: An Algorithm for Self-Assembly of Power-Delivery Networks using Recursive Branching in Information-Limited Environments*
}

\author{{Mohammadali Rashidioun}$^{1}$ and {Sangwoo Park}$^{1}$ and {Petras Swissler}$^{1}$%
\thanks{*This work was supported by {the NJIT Grace Hopper AI Research Institute}}%
\thanks{$^{1}$All authors are with {the department of  Mechanical and Industrial Engineering, New Jersey Institute of Technology, Newark, NJ 07102, USA}
        {\tt\small {mr943@njit.edu; sangwoo.park@njit.edu; petras.swissler@njit.edu}}}%
}
\usepackage[ruled,vlined]{algorithm2e}

\usepackage{tikz}

\newcommand\copyrighttext{%
  \footnotesize © 2026 IEEE. Personal use of this material is permitted. Permission from IEEE must be obtained for all other uses, in any current or future media, including reprinting/republishing this material for advertising or promotional purposes, creating new collective works, for resale or redistribution to servers or lists, or reuse of any copyrighted component of this work in other works.}

\newcommand\copyrightnotice{%
  \begin{tikzpicture}[remember picture,overlay]
    \node[anchor=south,yshift=10pt] at (current page.south) {\fbox{\parbox{\dimexpr\textwidth-\fboxsep-\fboxrule\relax}{\copyrighttext}}};
  \end{tikzpicture}%
}

\begin{document}

\maketitle
\copyrightnotice
\thispagestyle{empty}
\pagestyle{empty}

\begin{abstract}
Delivering sustained power to distributed equipment in unstructured field environments using pre-planned wired networks or battery-based solutions presents significant infrastructure and logistics 
challenges.
This paper presents Dendritic Recursive Pivoting (DeRP), a decentralized framework for multi-target network formation in robot swarms based solely on local communication and bearing-based sensing toward sinks.
We envision a system in which robots, acting as a conduit, self-assemble a power network from a common source, forming branches at locally selected pivot points that approximate the Steiner points of Steiner trees to efficiently route to multiple Sinks. This branching operation is performed recursively to enable scalable and adaptive network formation without global planning. 
The proposed method is evaluated in terms of the total network length and estimated power loss, and is quantitatively compared against global baselines such as the Minimum Spanning Tree and Steiner tree solutions (GeoSteiner), which require complete knowledge of Sink locations. Specifically, we found that the networks formed by DeRP asymptotically form approximately 125\% of the global minimum length while reducing power losses to 65\% relative to Euclidean Steiner trees. In addition, we empirically characterize scaling behavior by measuring simulation completion time as the number of Sinks and robots increases, and find that this scaling was sub-linear for up to 100 sinks. The proposed approach enables resilient, adaptive power delivery in environments where deployment  of traditional infrastructure is challenging.
\end{abstract}

\section{Introduction}

Power delivery is a major challenge of operating robotic equipment in the field. While batteries and gas generators can provide remote power sources, their bulk and inherent limits to endurance mean that these solutions are only effective for relatively short-duration deployments\cite{aubin2022towards}. For more permanent deployments, delivery via a network of power cables can provide a more efficient and reliable approach to power delivery, but such an approach requires a significant upfront investment in infrastructure\cite{RASTGOU2024114062}. In this paper, we present an approach for robots to self-assemble an on-demand wired power network via an efficient, recursive algorithm to deliver power from a centralized power ``Source'' to an arbitrary number of demand ``Sinks.''

The natural world provides compelling evidence that near-optimal multi-terminal networks can emerge from local rules. 
In both living and non-living systems, dendrites grow based on environmental gradients into complex and optimized networks through the process of dendritic arborization \cite{arikkath2012molecular, luczak2006spatial, kumar2022theoretical}.
In living systems, the slime mold \textit{Physarum polycephalum} is a particularly interesting  example: without centralized control, it grows transport networks that rival human-engineered systems in efficiency \cite{tero2010rules, ray2019information}.
A similar principle governs leaf venation, where new vascular
tissue extends incrementally toward the nearest under-served
region guided only by local auxin concentration
gradients \cite{runions2005modeling}. 
Ronellenfitsch and Katifori \cite{ronellenfitsch2016global} showed that this
coupling between incremental outward growth and local adaptive
feedback is what allows biological networks to
escape local minima and move toward dramatically improved topologies.%

\begin{figure}
    \centering
    \includegraphics[trim=0mm 0mm 0mm 0mm, clip, width=1.0\columnwidth]{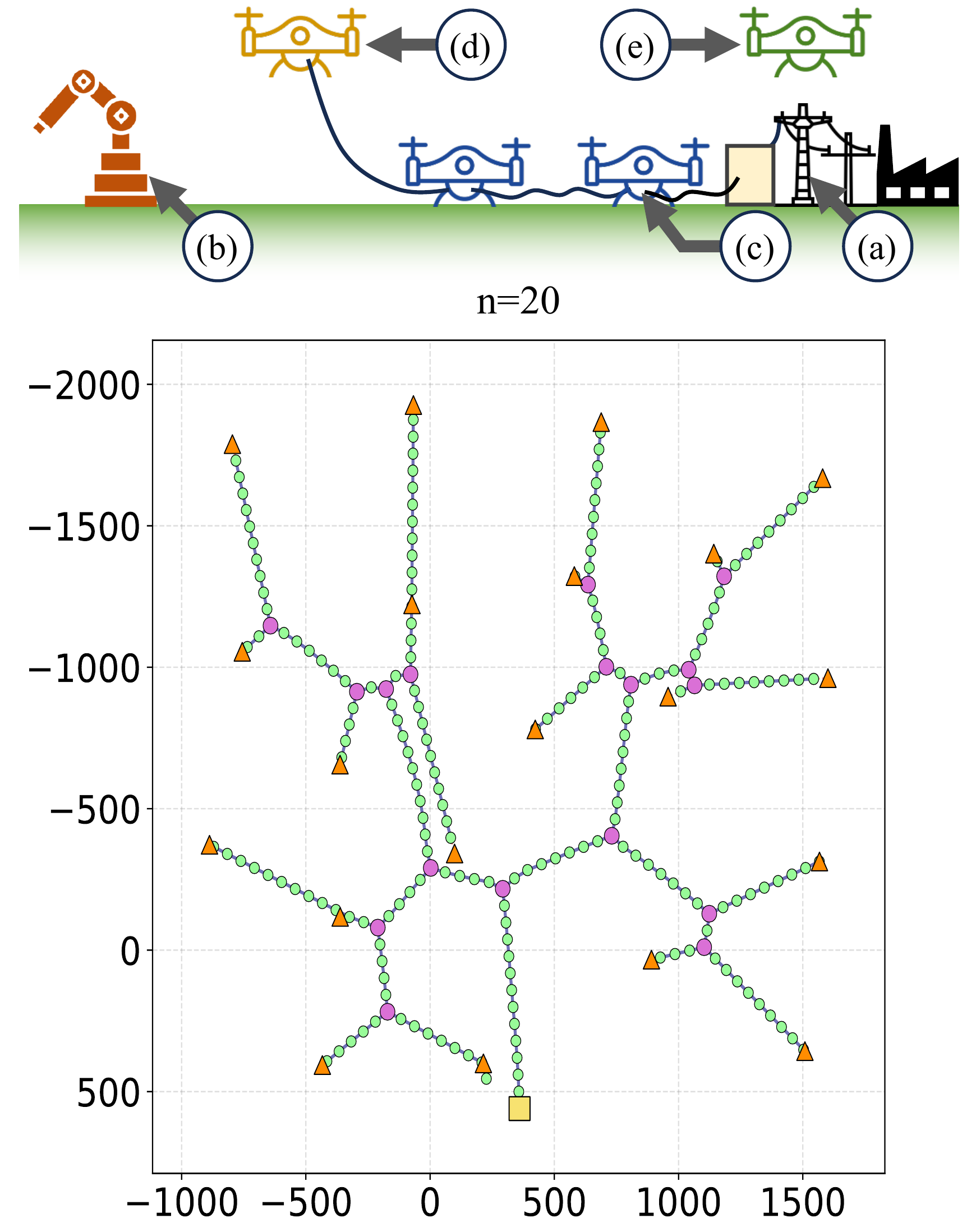}
    \caption{(Top) Concept for a robotic system able to create on-demand power networks from an easy-to-access power ``Source'' (a) to one or more remote, power-hungry devices, referred to in this paper as ``Sinks'' (b). Drones with deployable extension cords connect to this power Source, then fly some distance away and land on the ground (c). Other such drones (d) will, in turn, connect to these robots and extend the power network. Additional robots are released from the Source to build the network until it is completed (e). (Bottom) An efficient network formed via recursive branching.}
    \label{fig:concept_platform}
\end{figure}

Robot swarms have demonstrated analogous self-organizing
capabilities in path and chain formation. Nouyan et al. \cite{nouyan2008path}
showed that a swarm of simple robots can form a connected
physical chain between a Source and a single target using only local communication, without
any global map or centralized planner. More recent work has extended this to constrained platforms such as nano-drones operating in open spaces \cite{barcis2022chain}, and to relay deployment
problems where robots maintain communication links in
multi-robot teams \cite{pei2012steiner}. \textbf{However, these existing approaches are limited to single-path connectivity}: they
connect one Source to one target, but do not address the problem of self-assembling a branching tree that efficiently routes from one Source to multiple terminals simultaneously.
The problem of connecting a set of terminals with 
minimum total edge length in the Euclidean plane is 
formalized as the Euclidean Steiner Minimum Tree 
(ESMT) problem — a classic NP-hard combinatorial 
optimization problem~\cite{garey1979computers}.
Unlike the Minimum Spanning Tree (MST), which connects terminals by edges between them directly, the Steiner tree may introduce additional intermediate junction points called Steiner points that reduce the total length.
On the algorithmic side, centralized solvers such as
GeoSteiner \cite{juhl2018geosteiner} yield globally optimal
Steiner trees but require a priori knowledge of all Sink locations and offline computation before deployment,
conditions that are frequently unavailable in the field.
Decentralized approaches for multi-target deployment have
been proposed \cite{majcherczyk2018decentralized}, but these
optimize for connectivity preservation rather than network
length and do not compare against optimal routing baselines.

The problem we address in this paper sits at the intersection of these
two bodies of work, in the domain of functional
self-assembly, where swarms physically construct to form structures \cite{werfel2014designing,
rubenstein2014programmable, swissler2024behavioral}. 
Like biological transport networks, the structure
must branch efficiently toward multiple targets; like robotic
chain formation, it must emerge from bottom-up interactions
of simple agents. Unlike either, it must simultaneously
satisfy the constraints of physical robot deployment while
achieving a total network length competitive with globally
optimal baselines: the Minimum Spanning Tree (MST) and
the Steiner Minimum Tree (SMT).
Inspired by dendrite arborization, we consider a
swarm of mobile robots tasked with forming a physical power
delivery network from a single Source to $n$ spatially
distributed Sinks using limited communication and
sensing, with no global knowledge of Sink
positions.

This work has two major contributions. First, to our knowledge, the use of swarms and self-assembly approaches to power delivery networks is a novel application that has not previously been explored. Second, we present an algorithm for achieving these power networks via a recursive process of linear growth and pivoting, which we name Dendritic Recursive Pivoting (DeRP).

\section{Conceptualized Robot Platform}

As the concept of a self-assembling power network has not previously been considered in the literature, it is necessary to conceptualize what a hardware platform capable of performing this task would look like in order to ground our algorithm in current technological capabilities, with the understanding that significant work would be necessary to actually develop and deploy a real-world system. Fig. \ref{fig:concept_platform} shows a conceptual illustration of such a system. We anticipate such a system finding use in defense and disaster applications, where fast deployment is necessary, the ability to globally communicate may be limited, and where a priori knowledge is unavailable. We envision such a system to be used at large scales (e.g. $>$\qty{100}{km^2}).

DeRP assumes four capabilities of the robots:
    \paragraph{Movement and connection} Robots are able to attach a power cable to some fixed position and then extend the network by moving to some other location while deploying a power cable. Although conceptualized with flying robots, this could also be done with ground vehicles.
    
    \paragraph{Local communication} Robots are able to communicate within some communication radius $r_\mathrm{comm}$, such as via Bluetooth Low Energy, IR, or Li-Fi.

    \paragraph{Relative neighbor pose estimation} Robots can
    estimate the relative position and bearing towards neighboring robots within communication range.
    This capability is 
    consistent with the directed sensing model assumed in hierarchical
    bearing-based frameworks \cite{zhang2024hierarchical}, where each robot tracks the bearing to its Parent neighbors using only onboard sensors and without access to a global reference frame.

    \paragraph{Sensing heading towards Sinks} Robots can sense the
    direction toward Sinks. Robots do not require knowledge of their distance to or global position of these Sinks.
    This can be realized across operationally relevant scales by equipping Sinks with active RF or IR beacons. At short range, circumferential IR receiver arrays provide a lightweight bearing estimate within typical communication distances, without demanding distance measurements. 
    At field scale, UAV-based radio direction-finding 
    systems have demonstrated bearing estimation toward active VHF beacon transmitters at ranges exceeding one 
    kilometer \cite{cliff2015uavrt}, with median bearing errors below $7\degree$. Similarly, RF bearing toward active emergency locator beacons is common in search and rescue operations at distances of several kilometers \cite{cospas2020}. Within the swarm robotics literature, bearing-based homing toward active beacons has been demonstrated on both ground and aerial platforms \cite{decroon2019sgba}, confirming it as a well-established sensing primitive.

\section{DeRP Algorithm Description}
\label{sec:Algorithm_description}

\begin{figure}[t]
    \centering
    \includegraphics[width=\columnwidth]{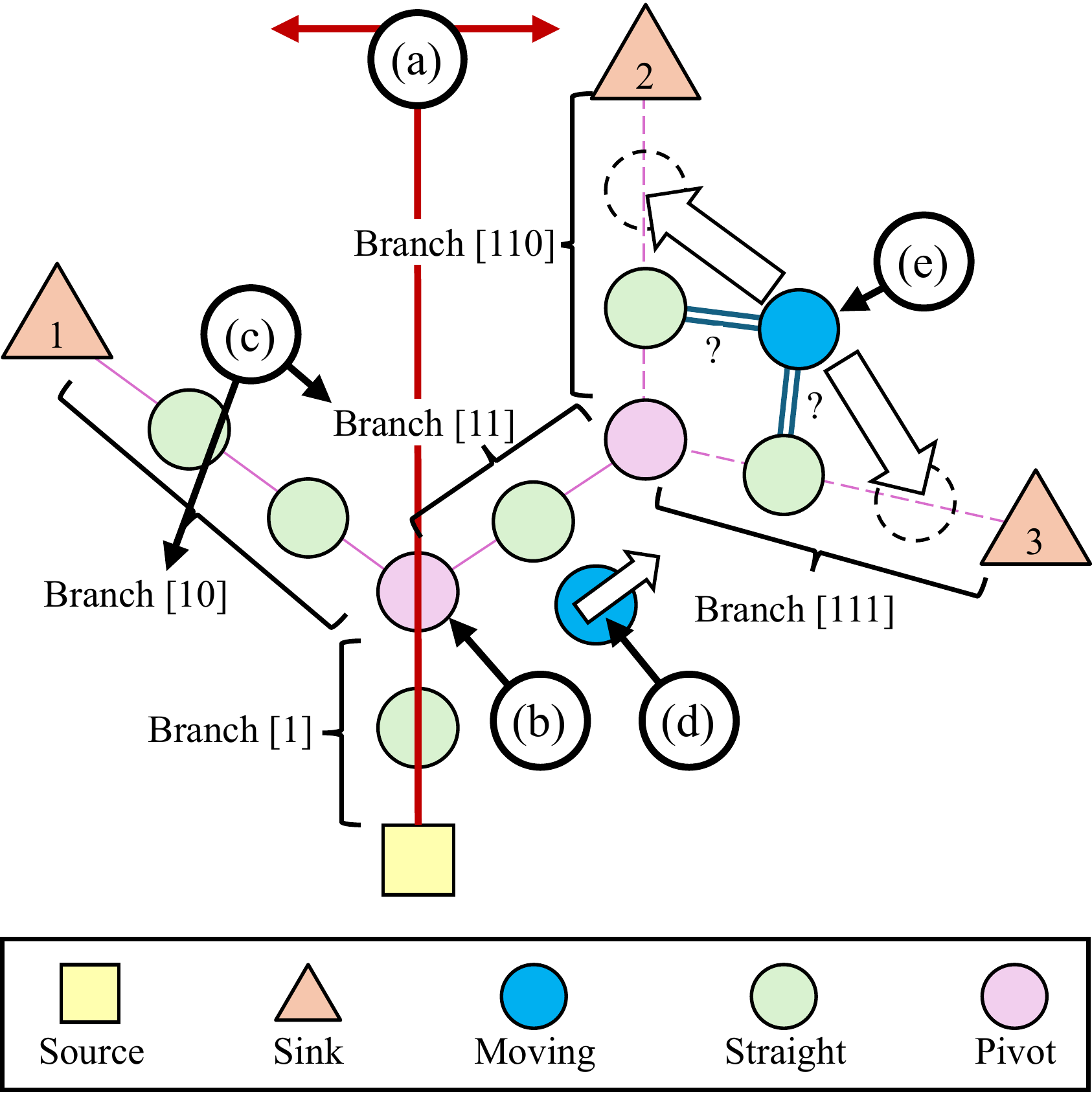}
    
    \caption{Illustration of the key algorithmic mechanisms. 
(a) Each Pivot node partitions its local environment into two half-planes using the incoming branch direction as a reference axis: starboard ($+$, right) and port ($-$, left). 
(b) The robot with the lowest $E(r)$ metric (closest to the ideal $120\degree$ Steiner angle with respect to its starboard and 
port Sink sets) is promoted to Pivot among its local 
neighborhood. (c) Upon branching, the ``Pivot'' appends \texttt{"1"} to the branch ID of its starboard Child and \texttt{"0"} to the branch ID of its port Child, propagating unique identifiers 
recursively along each branch. 
(d) Moving Robots without an active recruitment signal follow 
the guidance direction broadcast by nearby Network Robots. (e) When a Moving Robot receives simultaneous guidance messages from multiple Network Robots, it prefers to follow the robot with the highest branch ID. This deterministic priority rule prevents oscillation between competing recruiters, ensuring robots commit to a single branch.}
    \label{fig:branch_id}
\end{figure}

\begin{figure}[t]
    \centering
    \includegraphics[width=0.7\columnwidth]{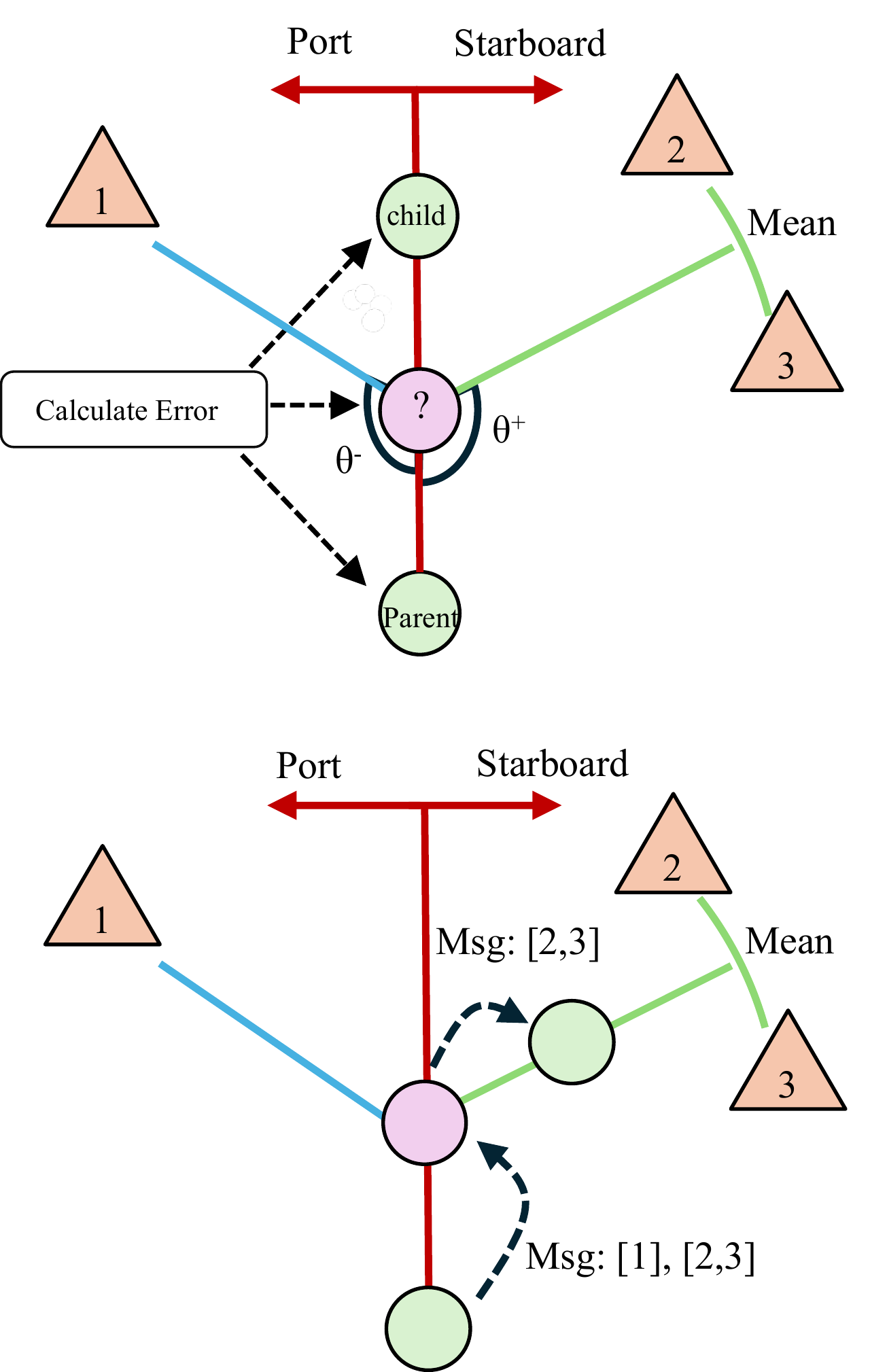}

    \caption{Branch selection process. (Top) Message aggregation and Error evaluation using $E(r)$. 
    (Bottom) Updated branching configuration.}
    \label{fig:branching_process}
\end{figure}

DeRP enables a swarm of homogeneous robots to self-assemble into a network that connects a fixed Source to an arbitrary number of spatially distributed Sinks. Starting from the Source, the network grows incrementally as robots are recruited one by one into a tree structure, branching recursively at locally selected pivot nodes to route toward multiple Sinks simultaneously. This branching process repeats at each level of the tree until every Sink has been reached, yielding a fully connected network without global planning or exact knowledge of Sink locations: each robot acts solely on bearing measurements to nearby Sinks and short integer messages exchanged with immediate neighbors.
The quality of the emergent network is evaluated in section~\ref{sec:Algorithm_validation}.

\subsection{Robot Roles}
Robots take on one of two roles that determine their behaviors: Moving or Network Robot.

 \paragraph{Moving Robots}
All robots begin in the \textit{Moving} role and navigate 
autonomously according to a three-priority hierarchy 
(Algorithm~\ref{alg:moving}): 
\begin{itemize}
    \item Respond to a Network Robot 
    requesting a ``Child.''
    \item Follow the growth direction broadcast by a Network Robot that is not requesting a Child (due to already reaching its Child limit). When multiple Network Robots provide guidance simultaneously, the Moving Robot 
    preferentially follows the one with the highest branch 
    ID value. This is how Moving Robots navigate without global information
    \item Drift toward the Source otherwise.
\end{itemize}

This ordering ensures that 
direct attachment opportunities always override passive 
navigation. Upon being accepted by a Network Robot as a 
Child, a Moving Robot transitions to the 
\textit{Network} agent (Algorithm~\ref{alg:moving}).

\begin{algorithm}[t]
\caption{Moving Robot Behavior}
\label{alg:moving}
\SetAlgoLined
\DontPrintSemicolon

\textbf{initialize} role $\leftarrow$ Moving\;

\While{role $=$ Moving}{
    Receive messages from all robots $j$ within $r_{\mathrm{comm}}$
    
    \If{a Network Robot requests a Child}{
        Among requesting Network Robots, move toward the nearest one\;
    }
    \Else{
       Select the nearby Network Robot with the highest 
        branch ID that provides guidance\;
        \If{a guiding Network Robot is found}{
            Follow the Network growth direction from Parent to Child\;
        }
        \Else{
            Move toward the Source\;
        }
    }
    
    \If{accepted by a Network Robot}{
        role $\leftarrow$ Network\;
    }
}
\end{algorithm}

\paragraph{Network Robots}
Once a Moving Robot is recruited as a Child to  a Network Robot, it transitions to the 
Network role and begins participating in the tree structure 
as follows (Algorithm~\ref{alg:network}).
Every Network Robot maintains a fixed spacing $R$ from its ``Parent'' in an assigned direction $\hat{d}$, using a local spacing controller (Eq.~1) to hold its position in the tree.
The controller computes a velocity command from two error terms:
\begin{align}
    \mathbf{v} &= K_R \, e_r \, \hat{u} \;-\; K_\theta \, 
        e_\theta \, \hat{u}_\perp \\
    e_r &= \|\mathbf{p}_{\mathrm{parent}} - \mathbf{p}_r\| - R \\
    e_\theta &= \hat u \times \hat u^* \;\triangleq\; 
        u_x u_y^* - u_y u_x^*
\end{align}
\noindent where $\hat u$ is the \emph{actual} unit vector from the 
robot toward its Parent; $\hat u^* = -\hat d$ is the \emph{desired} 
direction from the robot toward its Parent, implied by the assigned 
growth direction $\hat d$ (which points from Parent to Child); 
$\hat u_\perp$ is $\hat u$ rotated $90^\circ$ counter-clockwise; the 
cross product above denotes the scalar (perp-dot) product in the 
plane, equal to $\sin\theta$ for the signed angle $\theta$ from 
$\hat u$ to $\hat u^*$; and $K_R$, $K_\theta$ are proportional gains.

Simultaneously, it actively recruits a Child by broadcasting a request to nearby Moving agents within its communication range. The recruitment behavior differs by the node subtype:

A \textit{Straight} node broadcasts a generic 
    Child request and, upon accepting a Child, passes its 
    full set of assigned Sinks $\mathcal{S}$ and branch 
    ID downstream. The growth direction $\hat{d}$ passed 
    to the Child is computed as the normalized sum of 
    unit vectors toward each assigned Sink, i.e., a mean-bearing rule.
    This treat each Sink as equally weighted regardless 
    of distance, ensuring that the growth direction 
    reflects the angular distribution of Sinks rather 
    than being biased toward nearby ones.
    
A \textit{Pivot} node partitions its local 
    environment using the incoming branch direction 
    $\hat{u}_{\mathrm{in}} = \frac{\mathbf{x}_i - 
    \mathbf{x}_{\mathrm{Parent}}}{\|\mathbf{x}_i - 
    \mathbf{x}_{\mathrm{Parent}}\|}$ as a reference axis. 
    The line passing through the Pivot perpendicular to 
    $\hat{u}_{\mathrm{in}}$ divides the surrounding space 
    into two half-planes: starboard ($+$) and port ($-$). 
    A Moving Robot determines its side by evaluating 
    $\mathrm{sign}\left(\hat{u}_{\mathrm{in}} \times 
    \hat{v}_{ic}\right)$, where $\hat{v}_{ic}$ is the 
    unit vector from the Pivot to the Moving Robot. 
    Only robots on a needed side respond to the request. 
    Each Child is then assigned exclusively to one side: 
    the starboard Child receives $\mathcal{S}^+$ and 
    growth direction $\hat{d}^+$, and the port Child 
    receives $\mathcal{S}^-$ and growth direction 
    $\hat{d}^-$.

\begin{algorithm}[t]
\caption{Network Robot Behavior}
\label{alg:network}
\SetAlgoLined
\DontPrintSemicolon

\While{role $=$ Network}{
    
    Maintain spacing $R$ from Parent in direction $\hat d$\;
    
   Receive messages from all robots $j$ within $r_{\mathrm{comm}}$

    \If{recruitment is allowed}{
        
        \If{node type = STRAIGHT}{
            Broadcast Child request\;
        }
        \Else{ 
            Compute incoming direction 
            $\hat u_{\text{in}} = \frac{x_r - x_{\text{Parent}}}{\|x_r - x_{\text{Parent}}\|}$\;
            Broadcast side-specific Child request (STAR / PORT)\;
        }
    }
    
    \ForEach{reply from a free robot}{
        
        \If{recruitment is still allowed}{
            
            \If{node type = STRAIGHT}{
                Assign all Sinks and branch label to Child\;
            }
            \Else{ 
                Compute Child direction 
                $\hat u_c = \frac{x_c - x_r}{\|x_c - x_r\|}$\;
                Determine side: $\sigma \leftarrow \text{sign}(\mathrm{cross}(\hat{u}_{\mathrm{in}}, \hat{u}_c))$\;
                \If{side $\sigma$ is already occupied}{
                  Reject this robot\;
                  \textbf{continue}\;
                }
                
            }
            
            Set Child growth direction using mean-bearing rule \;
            Attach Child and notify it\;
            \textbf{break}\;
        }
    }
}
\end{algorithm}

\subsection{Pivot Selection and Branch Formation}
\label{sec:pivot}

The core novelty of the algorithm lies in how the Network Robots 
locally decide to branch. In an optimal Steiner tree, every 
junction or Steiner point satisfies a necessary geometric condition: 
the three edges meeting at each Steiner point form pairwise 
angles of exactly $120\degree$ \cite{hwang1992steiner}. Our 
algorithm approximates this condition locally by having each 
settled Network Robot evaluate how closely its own geometry 
resembles a Steiner point.

Specifically, each Network Robot $r$ with Parent $p$ and 
Child $c$ constructs three directional axes: one from $r$ 
towards $p$, one toward the average bearing of the starboard Sinks 
$\mathcal{S}^+$, and one toward the average bearing of the port 
Sinks $\mathcal{S}^-$. The pivot eligibility metric $E(r)$ 
measures the total angular deviation of these three axes 
from the ideal $120\degree$ configuration:

\begin{equation}
    E(r) = |\theta^+ - 120\degree| + |\theta^- - 120\degree|
\end{equation}
\noindent where $\theta^+$ and $\theta^-$ are the angles between 
$\hat{u}_{p \to r}$ and the average bearing toward 
$\mathcal{S}^+$ and $\mathcal{S}^-$ respectively. A 
smaller $E(r)$ indicates a geometry closer to the 
optimal Steiner configuration.

Pivot selection relies solely on local communication; no global knowledge of Sink positions, network topology, or the eligibility of other robots is required. After its Child has settled into its desired position at spacing $R$, each Network Robot compares its own metric against those of its immediate Parent and Child only. If $E(r) < E(p)$ 
and $E(r) < E(c)$, robot $r$ promotes itself to a Pivot 
node and initiates a branch split toward starboard and 
port (Algorithm~\ref{alg:pivot}).

\begin{algorithm}[t]
\caption{Pivot Selection}
\label{alg:pivot}
\SetAlgoLined
\DontPrintSemicolon

\KwIn{Network Robot $r$ with Parent $p$, Child $c$, and assigned Sinks $\mathcal{S}_r$}
\textbf{Require:}{$|\mathcal{S}_r| \geq 2$}

Split $\mathcal{S}_r$ into $\mathcal{S}_L$, $\mathcal{S}_R$ on either side of $\hat{u}_{p \to r}$\;

\If{$\mathcal{S}_L = \emptyset$ \textbf{or} $\mathcal{S}_R = \emptyset$}{
    \Return \tcp*{all Sinks on one side, no split needed}
}

Compute pivot metric:
$E(r) = |\theta_L - 120\degree| + |\theta_R - 120\degree|$

where $\theta_L$, $\theta_R$ are angles between $\hat{u}_{p \to r}$ and the average bearing toward $\mathcal{S}_L$, $\mathcal{S}_R$\;

\If{$E(r) < E(p)$ \textbf{and} $E(r) < E(c)$}{
    $r$ becomes a pivot node and initiates branch split\;
}

\end{algorithm}

\subsection{Termination Condition}

A branch terminates when the last robot in the branch comes within 
$r_{\mathrm{Sink}}$ of its assigned Sink. Upon touching 
a Sink, the end of the branch initiates a completion flag that travels upstream through the chain until it 
reaches the nearest Pivot node, signaling that no further 
guidance should be broadcast on that branch. The full 
network is complete when all $n$ Sinks have been connected to the network.

\section{Algorithm Validation}
\begin{figure}[t]
    \centering
    \includegraphics[trim = 0mm 0mm 0mm 0mm, clip, width=\columnwidth]{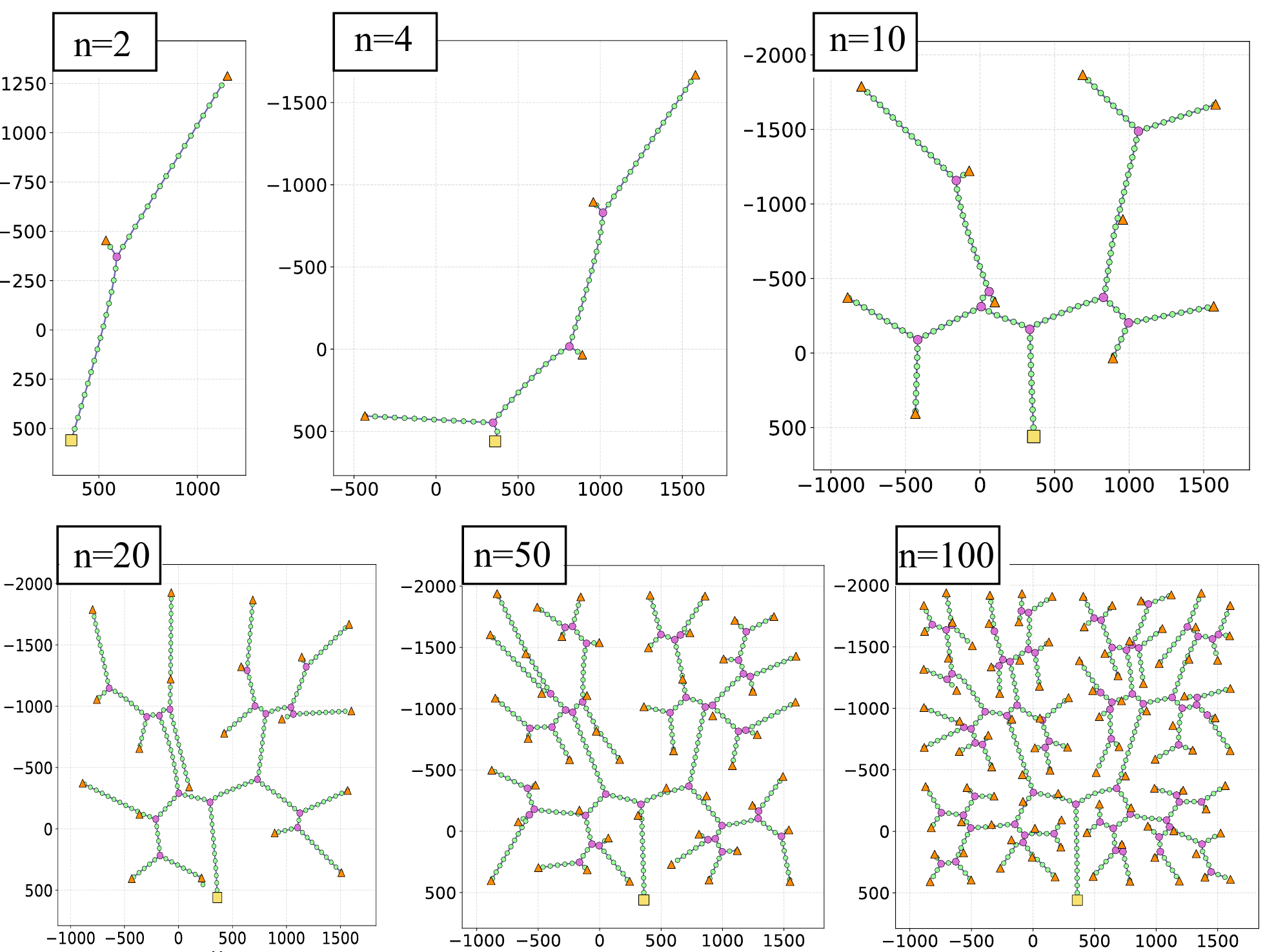}
   
    \caption{DeRP network topology for $n \in \{2, 4, 10, 20, 50, 100\}$ 
Sinks. Branching depth and number of pivots grow with $n$.}
    \label{fig:tiles}
\end{figure}

\begin{figure}
    \centering
    \includegraphics[trim=0mm 0mm 0mm 0mm, clip, width=0.75\linewidth]{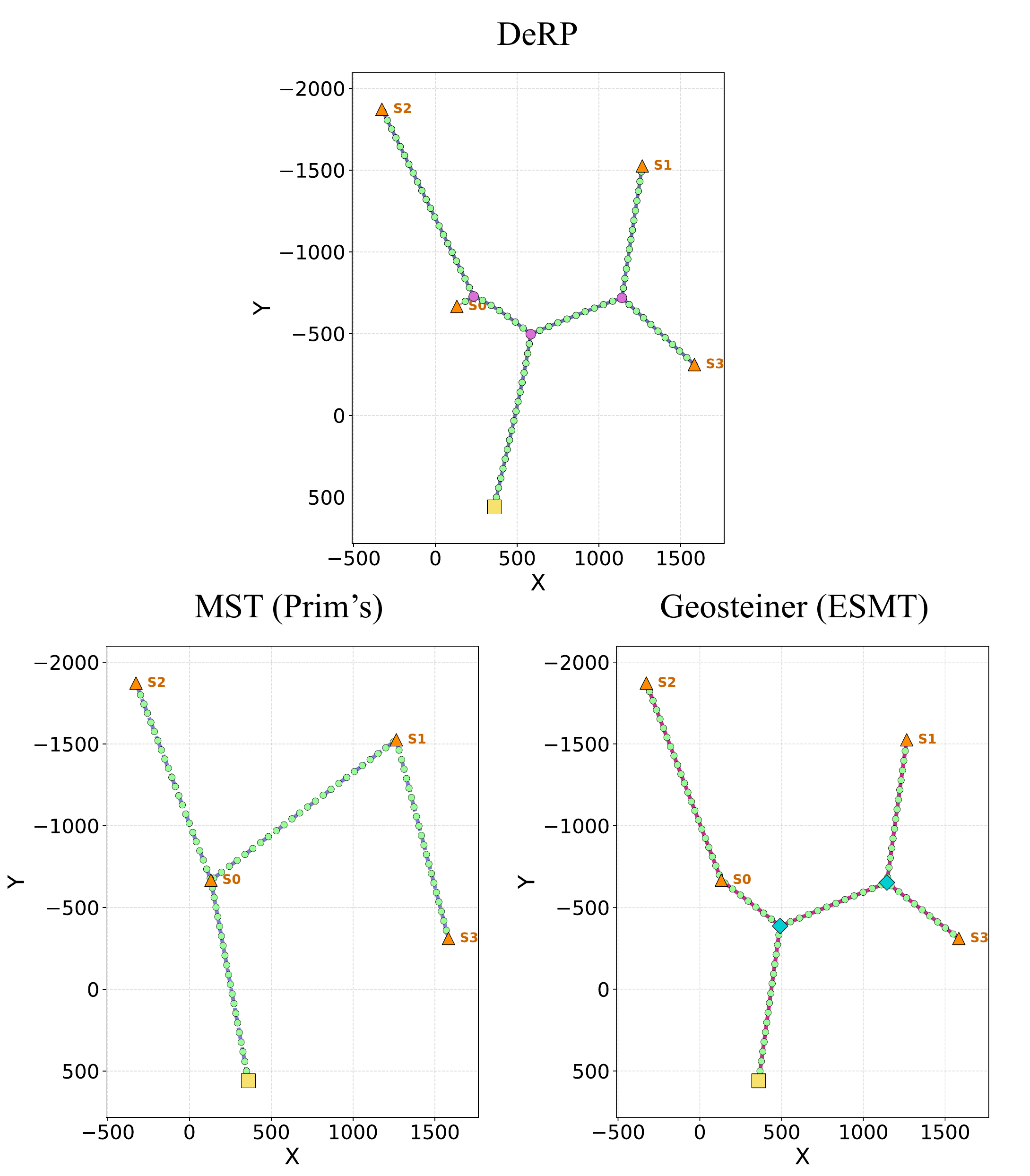}

  \caption{{Quantitative Metric Comparison of Network Topologies:}
(Top) DeRP; (Bot. Left) MST; (Bot. Right) GeoSteiner- ESMT. }
    \label{fig:test_scenario}
\end{figure}

To investigate how the proposed approach scales with problem size, we vary the number of Sinks with quantities $n \in \{2, 4, 6, 8, 10, 12, 15, 20, 30, 40, 50, 100\}$:
a total of 12 configurations. 
\newline
In each trial, Sinks are 
placed uniformly at random within a fixed 2500 $\times$ 2500 
unit arena, subject to a minimum pairwise separation of 
$r_{\mathrm{comm}}$ to avoid clustered configurations. The Source is fixed at a constant position 
across all trials.

Robots are introduced into the arena sequentially at a fixed spawn interval beginning at the Source location. Each robot starts in the Moving role and follows the behavioral rules described in   Section~\ref{sec:Algorithm_description} until it is recruited into the network. No upper limit on total robot count is imposed during a trial; the simulation terminates when all Sinks have been reached. All other parameters are held constant across scenarios at the values reported in Table~\ref{tab:params}.
Fig.~\ref{fig:tiles} illustrates representative completed networks 
for $n \in \{2, 4, 10, 20, 50, 100\}$. At small Sink counts, the 
network forms a shallow tree with few Pivot nodes; as $n$ grows, 
the recursive branching structure produces progressively deeper 
trees that spread to cover the arena. This qualitative progression 
is consistent with the quantitative trends reported in the 
following subsections.

\label{sec:Algorithm_validation}
\begin{table}[t]
\centering
\caption{Simulation Parameters}
\label{tab:params}
\renewcommand{\arraystretch}{1.15}
\begin{tabular}{lcc}
\hline
\textbf{Parameter} & \textbf{Symbol} & \textbf{Value} \\
\hline
Inter-robot spacing     & $R$                  & 60 units  \\
Communication range     & $r_{\mathrm{comm}}$  & 180 units \\
Arena size              & —                    & $2500 \times 2500$ units \\
Sink touch distance     & $r_{\mathrm{Sink}}$  & 60 units  \\

Minimum Sink separation & —                    & 180 units  \\
Number of Sinks         & $n$                  & 2--100    \\
Trials per configuration & —                  & 50    \\
\hline
\end{tabular}
\end{table}

We evaluate DeRP based on three efficiency metrics:

\paragraph{Network Length.} The primary structural metric is the 
total length of the assembled robot network, defined as the 
sum of all Parent-Child edge lengths in the tree. To assess 
quality relative to optimal baselines, we report the ratio 
of robot network length to the optimal Steiner Tree length computed over the Source and all Sinks using GeoSteiner \cite{juhl2018geosteiner}, which represents the tightest possible lower bound. A ratio near 1.0 indicates near-optimal routing. We also compare against a Minimum Spanning Tree length.

\paragraph{Completion timesteps} We measure empirical complexity 
as the total number of fixed-timestep simulation steps 
required to reach all Sinks, recorded as a function of 
Sink count $n$. Since robots are spawned at a fixed rate throughout 
the simulation, completion time reflects both the geometric 
difficulty of the problem and the recruitment dynamics of 
the algorithm. This allows us to characterize the observed 
scaling behavior as the number of Sinks grows.

\begin{figure*}[]
    \centering
    \includegraphics[trim=0mm 2mm 0mm 0mm, clip, width=\standardreducedwidth\linewidth]{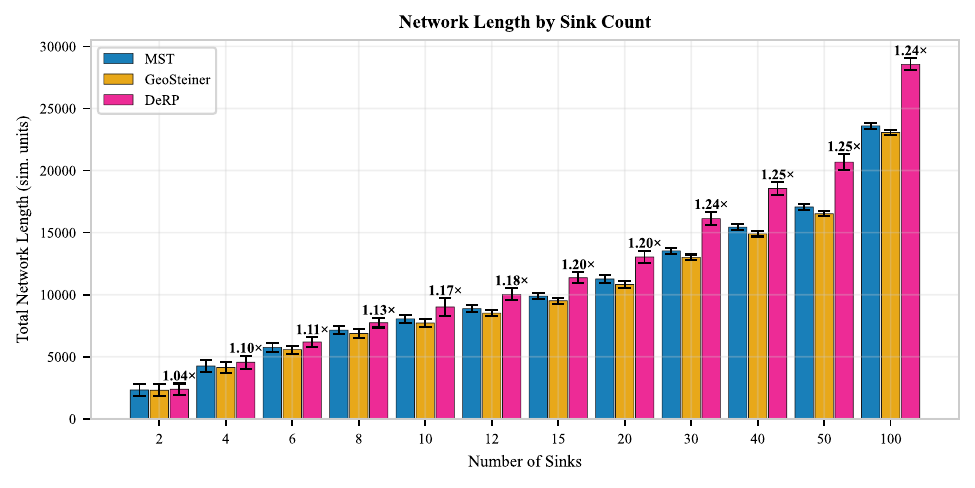}
    \vspace{-3mm}
    
    \caption{Total network length comparison across Sink counts. 
Error bars indicate $\pm 1\sigma$ over all trials
    The DeRP method is compared against the Minimum Spanning Tree (MST) 
    and the geometric Steiner tree (GeoSteiner) baseline. 
    Error bars indicate standard deviation over 50 trials. 
  Annotation on the DeRP bar indicates the ratio of DeRP length to the GeoSteiner optimal baseline.}
    \label{fig:network_length}
\end{figure*}

\paragraph{Power Loss}
To evaluate energy efficiency, we model power transmission loss 
using the standard $I^2R$ distribution network analogy \cite{Silva2023QUBO, Borisov2022}. 
Since robots maintain a constant inter-robot spacing of $R = 60$, 
the number of robots on each branch serves as a direct proxy for 
branch resistance ($R_e \propto n_{robots}(e)$), and the current 
$I_e$ through each branch is proportional to the
fraction of Sinks served downstream, relative to the total Sink
count, reflecting aggregated power demand. Total network 
power loss is then:
\begin{equation}
P_{total} = \sum_{e \in E} I_e^2 \cdot n_{robots}(e)
\label{eq:power}
\end{equation}
\noindent where $E$ is the set of network edges, and $I_e$ is defined as:
\begin{equation}
I_e = \frac{k_e}{n}
\label{eq:current}
\end{equation}
\noindent where $k_e$ is the number of Sinks downstream of 
branch $e$ and $n$ is the total Sink count. This normalization 
models each Sink as drawing an equal share of a fixed 
total current, isolating topological efficiency from absolute 
power demand.
Although GeoSteiner minimizes total network length, 
length minimization is a linear objective, whereas 
power loss, defined in Eq.~\eqref{eq:power}, 
penalizes current on shared segments quadratically. 
These two objectives are not equivalent, and 
minimizing one does not necessarily minimize the other.

\subsection{Network Length}
Although DeRP operates without global knowledge of Sink positions, the emergent network lengths are competitive with the centralized baselines. Fig.~\ref {fig:network_length} reports the mean total network length for DeRP, MST, and Geosteiner across all Sink counts. The ratio between our DeRP network length and the Geosteiner-provided optimal Steiner tree length- is annotated directly on each bar as a primary measure of routing quality. 
The DeRP length remains within  $1.04\times$--$1.25\times$ of the Geosteiner optimal across all configurations, with the ratio stabilizing beyond $n=30$ and showing no unbounded growth at $n=100$. The average network length per Sink decreases monotonically from $1193$ units at $n=2$ to 
$286$ units at $n=100$ for the robot network, a result of increasing path sharing as branching depth grows.

\subsection{Completion timesteps}

\begin{figure}
    \centering
    \includegraphics[width = \standardreducedwidth\columnwidth]{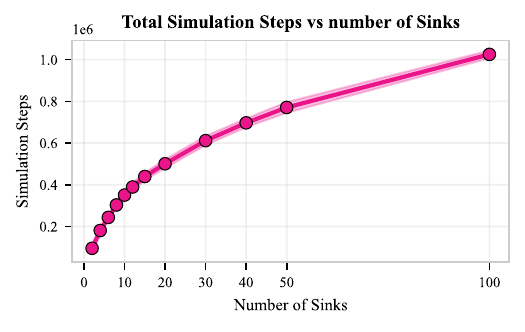}
    
    \vspace{-5mm}
    
    \caption{Total simulation steps as a function of the number of Sinks. }
    \label{fig:time}
\end{figure}

  \begin{figure*}
    \centering
    \includegraphics[width=\standardreducedwidth\linewidth]{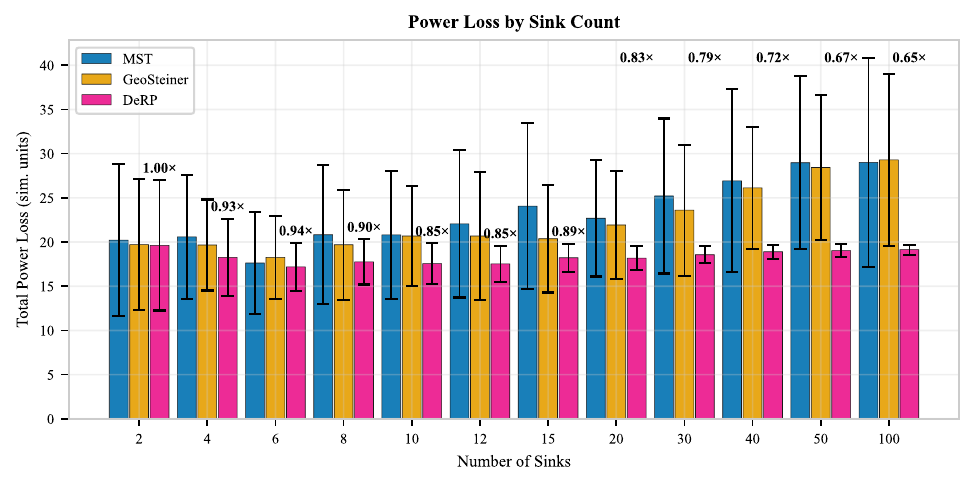}
    \vspace{-5mm}
    \caption{Power loss as a function of the number of Sinks for three routing strategies (MST, GeoSteiner, and Robot Network). Bars show the mean power loss over multiple runs and error bars indicate variability across trials (mean $\pm$ standard deviation). Annotation on the DeRP bar indicates the ratio of DeRP Power loss  to the GeoSteiner. }
    \label{fig:power_loss}
\end{figure*}
Fig.~\ref{fig:time} shows total simulation steps as a function 
of Sink count. Steps increase monotonically with problem size, yet sub-linearly: a $50\times$ increase in Sink count produces only a $10.8\times$ increase in total steps. 
This sub-linear trend is reflected in two complementary measures. 
First, the marginal step cost per additional Sink decreases from approximately $42{,}600$ steps at $n=2$--$4$ to $5{,}100$ steps beyond $n=50$, an $8\times$ reduction that reflects growing reuse of shared trunk segments as the network expands. Second, the mean number of robots deployed per Sink falls from $19.3$ at $n=2$ to $4.3$ at $n=100$, confirming that the network becomes structurally more efficient at scale. Both trends are consistent with the recursive branching structure of the algorithm: early trunk segments are shared by an increasing number of downstream branches, reducing the marginal cost of routing each additional Sink.

\subsection{Power Loss}

\begin{figure}[t]
    \centering
    \includegraphics[width=\standardreducedwidth\columnwidth]{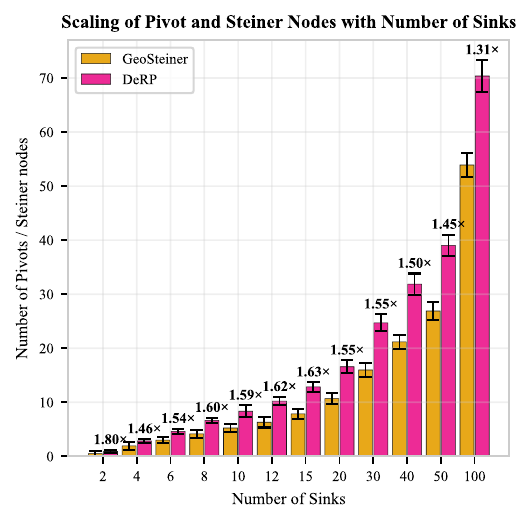}
    
    \vspace{-5mm}
    
    \caption{Pivot count versus number of Sinks for GeoSteiner and DeRP. Error bars denote standard deviation.}
    \label{fig:pivot_count}
\end{figure}

Fig.~\ref{fig:power_loss} compares power loss across all configurations. The most significant finding is that the DeRP power loss remains effectively constant across all tested Sinks. 
Minimizing power loss is an objective that quadratically penalizes high current on shared 
trunk lines. The DeRP method, by 
introducing branch points earlier, distributes 
current more aggressively and consistently 
outperforms GeoSteiner on power loss across all 
configurations. For example, at $n=50$, the robot network 
incurs only $68\%$ of GeoSteiner's power loss and 
$67.5\%$ of MST's, an advantage that grows 
steadily with problem size. This suggests 
that the local $120\degree$ branching criterion implicitly 
favors topologies that branch earlier, reducing 
the length of shared high-current segments near 
the Source and thereby lowering the total $I^2R$ loss.

\section{Discussion}
Across all three metrics, the DeRP algorithm produces 
networks that are competitive with centralized baselines 
despite relying solely on local information.
Network length remains within 1.04--1.25$\times$ of 
the GeoSteiner optimal, with the ratio stabilizing beyond 
$n=30$, indicating that the overhead of local decision-making 
is bounded. \newline
The most significant finding concerns power loss. Despite 
being longer than GeoSteiner, the 
robot network incurs lower mean power loss in every configuration. 
This is explained by the quadratic nature of $I^2R$: earlier, 
more aggressive branching reduces current aggregation on shared 
trunk segments near the Source.
\newline
A factor that increases network length is the higher number of Pivot nodes produced by the decentralized algorithm. 
Additional branch points introduce extra segments, which can increase total path length relative to the Steiner optimum. 
We therefore analyze the Pivot count across all configurations to quantify this structural overhead.
Since the robot network forms a tree in which every Pivot has exactly two Children, giving it degree three: the theoretical upper bound on the number of Pivot nodes follows directly from the degree 
sum formula $k \leq n - 2$,
where $k$ is the number of Pivot nodes and $n$ is the total number of terminals (Source $+$ Sinks), independently of any optimality argument \cite{hwang1992steiner}. Empirically, as shown in Fig.~\ref{fig:pivot_count}, DeRP produces on average $1.53\times$ the number of Pivots as Geosteiner across all configurations. This is a cost of local decision-making, since robots do not know globally whether a branch point is truly optimal. However, this ratio remains stable across 
all tested problem sizes, indicating that the overhead scales proportionally rather than growing without bound, and network lengths remain competitive despite the additional pivots.

\section{Conclusion}
The DeRP framework demonstrates that a swarm of robots without full global knowledge can self-assemble power 
delivery networks competitive with centralized 
baselines. Unlike MST and GeoSteiner, which 
require complete prior knowledge of all Sink 
locations and are not designed for decentralized execution, DeRP operates with local information only. It trades a bounded length overhead of up 
to $25\%$ relative to the ESMT optimum for this 
capability, while outperforming both baselines 
on power efficiency across all tested 
configurations. Empirically, the length ratio 
remains stable with problem size, and completion 
timesteps scale sub-linearly with the number 
of Sinks.

Several directions remain open for extending the presented algorithm. First, the current pivots are static. A natural extension would allow 
pivot Network Robots to dynamically adjust their positions 
after placement, iteratively correcting toward the 
exact $120\degree$ angle used in Steiner trees.
Secondly, the current starboard-over-port branching is based on a fixed criterion that always prefers starboard over port.  Future work 
could replace this with a demand-aware criterion. 
For example, prioritizing branches toward Sinks 
with lower remaining battery levels, allowing the 
network to adapt its routing to the urgency of 
power demand at each terminal.
Third, while the present evaluation uses a uniformly random Sink distribution, real deployments are likely to involve heterogeneous configurations. A systematic analysis of algorithm performance under such distributions is planned.
Finally, physical validation on a hardware swarm 
platform remains an important next step to confirm 
that the algorithm's performance translates from 
simulation to real-world deployment conditions.

\section*{Acknowledgment}

We thank Amin Taghieh, Nidhi Sakpal, and Pranay KC for their input and contributions throughout the project, and especially Maria Angel Palacios Sarmiento for assistance in preliminary investigations.

\textit{AI Disclosure:} Some helper functions were generated using Claude Anthropic and were checked for correctness. All algorithmic design was performed entirely by the authors. Source code available on request; all AI-assisted functions are noted in comments.

\balance
\vspace{-6pt}


\end{document}